\documentclass{article}

\usepackage{arxiv}

\usepackage[utf8]{inputenc} 
\usepackage[T1]{fontenc}    
\usepackage{hyperref}       
\usepackage{url}            
\usepackage{booktabs}       
\usepackage{amsfonts}       
\usepackage{nicefrac}       
\usepackage{microtype}      
\usepackage{lipsum}
\usepackage{times}
\usepackage{latexsym}

\usepackage{graphicx}
\usepackage{float}
\usepackage{titlesec}
\usepackage{arydshln}
\title{Transformer-Based Models for Question Answering on \emph{COVID19}}

\author{
Hillary Ngai $^1$ $^2$ \and Yoona Park  $^1$ $^2$ \and John Chen  $^1$ $^2$\and
Mahboobeh Parsapoor (Mah Parsa)  $^1$ $^2$\\
$^1$Vector Institute for Artificial Intelligence\\
$^2$University of Toronto,
\\
\texttt{\{hngai, ypark, johnc, mahparsa\}@cs.toronto.edu}

}

\begin{document}
\maketitle

\begin{abstract}
In response to the Kaggle's COVID-19 Open Research Dataset (CORD-19) challenge, we have proposed three transformer-based question-answering systems using BERT, ALBERT, and T5 models. Since the CORD-19 dataset is unlabeled, we have evaluated the question-answering models' performance on two labeled questions answers datasets \textemdash CovidQA and CovidGQA. The BERT-based QA system achieved the highest F1 score (26.32), while the ALBERT-based QA system achieved the highest Exact Match (13.04). However, numerous challenges are associated with developing high-performance question-answering systems for the ongoing COVID-19 pandemic and future pandemics. At the end of this paper, we discuss these challenges and suggest potential solutions to address them. 
\end{abstract}

\section{Introduction} 
\textit{Question Answering} or \textbf{QA} systems may be useful for the medical research community to stay up-to-date when new literature is rapidly growing. However, due to the lack of labeled question-answer pairs, developing accurate QA systems is challenging. Thus, one solution is to fine-tune pre-trained transformer-based QA systems \footnote{\footnotesize{A typical transformer-based QA system uses a parallelizable architecture to efficiently find  answers (i.e., a segment of text) to a query.}} \cite{ishwari2019advances, lukovnikov2020pretrained, van_Aken_2019}.

In this paper, we  propose three QA systems developed for the Kaggle \textit{COVID-19 Open Research Dataset,} or \textbf{CORD-19} dataset \footnote{\footnotesize{They launched a competition to provide a chance for the NLP community to develop QA systems for medical experts to find answers to high-priority medical questions related to COVID-19}}. We built the QA systems after carefully reviewing various QA systems submitted to the CORD-19 Kaggle competition.  We have also presented our preliminary results obtained from evaluating the performance of three transformers: \textit{Bidirectional Encoder Representations from Transformers} or \textbf{BERT} \cite{devlin2019bert}, ALBERT, and \textit{Text-to-text transfer transformer} or \textbf{T5} model on two new QA datasets \textemdash CovidQA and CovidGQA. Finally, we discuss the challenges of developing a high-performance QA system for COVID-19-related research and suggest solutions to improve performance.
\section{A Review of QA Systems for the Kaggle CORD-19 Dataset}
More than 1,000 teams participated in the Kaggle CORD-19 dataset competition. They used various \textit{Natural Language Processing)} or\textbf{NLP} approaches, including BERT \cite{devlin2019bert}). Since most QA systems in the competition were developed based on BERT and \textit{Latent Dirichlet Allocation} or \textbf(LDA), the following subsections just review LDA-based QA systems and BERT-based QA systems submitted to the competition. 
\subsection{LDA-based QA Systems}
Using LDA to develop QA systems is considerably uncomplicated \cite{celikyilmaz2010lda,cui2014intelligent,chinaei2014topic,Ji2012QuestionanswerTM}. Thus, for the CORD-19 Kaggle competition, various LDA-based QA systems were developed. For example, one approach \footnote{\footnotesize{among 388 submitted kernels}} \cite{Wolffram} combined the \textbf{whoosh} search engine and used Jensen-Shannon distance to discover topics related to \textit{What do we know about COVID-19 risk factors?} and to find documents similar to those topics. Another interesting LDA-based QA \footnote{\footnotesize{It was received a lot of attention among other teams and was one of the competition's kernels that obtained the highest number of votes around 783. }} was proposed in \cite{Maksim}; It combined k-means clustering and LDA to cluster documents and discover topics from clusters to facilitate extracting articles from the CORD-19 dataset. 
In \cite{smith2020nonpharmaceutical}, a combination of LDA and Anserini (i.e., an open-source information retrieval approach) was proposed to develop a QA system that could find articles related to non-pharmaceutical intervention. The main aim of this work was to discover new interventions in specific environments by incorporating the context for each response in the search and guide policymakers to take appropriate actions to control spreading COVID-19 virus. 

\subsection{BERT-based QA systems for CORD-19}
Since one of BERT's applications is in QA, different variations of BERT \cite{sanh2019distilbert, liu2019roberta, lan2020albert, chan2019recurrent} have been used. For the competition, many teams used BERT to develop QA systems \cite{yang2018anserini, Sandyvarma}. For instance, \cite{yang2018anserini} used BERT to find relevant answers to keywords extracted from a question. The found solutions were ranked by \textit{Universal Sentence Encoder Semantic Similarity} or \textbf{USESS} then \textit{Bayesian Additive Regression Trees} or \textbf{BART} summarized the top results. Another team employed  
BERT as a semantic search engine to find answers. The QA system produced semantically meaningful sentence embedding on the paragraph extracted from the CORD-19 dataset and found five paragraphs and their corresponding papers' titles and abstracts. In  \cite{Sandyvarma}, the QA systems
were based on \textit{A little BERT} or \textbf{ALBERT} \cite{lan2020albert} to find answers for questions related to COVID-19. 

\section{Transformer-Based QA Systems}
For the competition, we aimed to develop a QA system using a high performance Transformer. To do so, we developed three QA systems using BERT-large, ALBERT-base and \textit{Text-to-text transfer transformer} or \textbf(T5)-large, pre-trained them on various QA datasets
and evaluated them on two labeled questions answers datasets \textemdash CovidQA, and CovidGQA as described below. 
\subsection{Datasets for Pre-training Transformers} 
\label{dataset}
We use various datasets to pre-train our QA systems: 1) \textbf{SQuAD v1.1} \cite{rajpurkar2016squad} \textemdash the \textit{Stanford Question Answering Dataset} containing 100k question-answer pairs on more than 500 articles; 2) \textbf{SNLI} \cite{bowman-etal-2015-large} \textemdash the \textit{ Stanford Natural Language Inference} corpus containing 570k human-written English sentence pairs manually labeled for balance classification with the labels entailment, contradiction and neutral; 3) \textbf{MultiNLI} \cite{N18-1101} \textemdash the \textit{ Multi-Genre Natural Language Inference} corpus containing 433k crowd-sourced sentence pairs with the same format as SNLI except it includes a more diverse range of text and a test set for cross-genre transfer evaluation; 4) \textbf{STS} \textemdash the \textit{Semantic Textual Similarity} benchmark is a careful selection of data from English STS shared tasks (2012-2017) comprising of 8.6k annotated examples of text from image captions, news headlines, and user forums; and 5) \textbf{BioASQ} \footnote{\footnotesize{http://bioasq.org/}} \textemdash the question-answering biomedical dataset consists of 1k questions with ``exact" and ``ideal" answers. We specifically use BioASQ factoid QA pairs, excluding yes/no or list QA pairs, because the factoid dataset has a similar structure as SQuAD v1.1 \cite{rajpurkar2016squad}. 
\subsection{Datasets for Evaluating Transformers}
After pre-training our models, we evaluate our QA systems on \textbf{CovidGQA} and \textbf{CovidQA}. The former \footnote{\footnotesize{an example of a question in the CovidQA dataset is: What is the incubation period of the virus?}} is a COVID-19 dataset created manually and encompasses 198 general question-text-answers related to COVID-19\footnote{\footnotesize{an example of a question in the CovidGQA dataset is: How can I protect myself from getting COVID-19?}}. The question-text has been extracted from medical websites, and medical \textit{subject-matter experts}or \textbf{SMEs} have provided answers. The latter is a COVID-19 question-answering dataset built by hand from knowledge gathered from the Kaggle CORD-19 dataset \cite{tang2020covidqa}. We merged the CovidQA dataset with the CORD-19 dataset to extract each article's relevant text to answer each question in the dataset. The final evaluation dataset contained 69 question-text-answer triplets. 

\subsection{BERT-large}
Our BERT-large QA system is developed using a pre-trained QA BERT-large-uncased model with whole word masking fune-tuned on SQuAD v1.1 \cite{rajpurkar2016squad}. The model contains 24 Transformer blocks, 1024 hidden layers, 16 self-attention heads adding up to 340M parameters in total.

\subsection{ALBERT-base}
The ALBERT-base QA system (Figure \ref{fig:Bert}) is forned using a pre-trained QA ALBERT-base-uncased model fine-tuned on SQuAD v1.1 \cite{rajpurkar2016squad}. The model contains 12 Transformer blocks, 768 hidden layers, 12 self-attention heads, adding up to 12M parameters in total. 
\subsection{T5-large}
The T5-large QA system (see Figure \ref{fig:T5}) is based on the T5 model that is a modern, massive multitask model trained by uniting many NLP tasks in a unified text-to-text framework \cite{raffel2019exploring}. By leveraging extensive pre-training and transfer learning, it has achieved state-of-the-art performance on a variety of NLP benchmark tasks, including the GLUE benchmark~\cite{wang2018glue}. Following work by \cite{roberts2020much}, which explores the task of generative closed-book question answering, we explore the efficacy of generating (rather than extracting) COVID-19 answers directly from an input question, without context. Unlike our preceding two approaches, the T5 model explores generation of answers to questions, without context. 
Using the pre-trained T5 model with 770M parameters released by \cite{raffel2019exploring}, we fine-tune for 25000 steps on an equal-proportions mixture of three QA tasks using the Natural Questions dataset, Trivia QA dataset and the train split of the COVID-19 QA dataset \cite{kwiatkowski2019natural, joshi2017triviaqa}. Only the queries are given as input and answers are generated using simple greedy decoding. Evaluation is then performed on the test split of the COVID-19 dataset. We emphasize that these results are not directly comparable to the other frameworks, as the model is faced with the challenging task of jointly localizing relevant information and then generating a coherent answer. The advantage of such a framework is that it is context-free, meaning that it requires the least data preparation and human intervention. 

\subsection{Evaluation of QA Systems}
\begin{figure}
    \centering
    \includegraphics[scale=0.5]{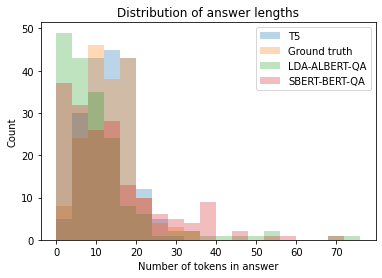}
    \caption{\footnotesize{Comparison of generated answer lengths on the CovidGQA dataset.}}
    \label{fig:t5_gt_answer_lengths}
\end{figure}

We evaluate the above transformers using two datasets \textemdash CovidQA and CovidGQA. Using the same evaluation metrics as SQuAD v1.1, we calculate the macro-averaged F1 score and \textit{Exact Match} or \textbf{EM} of the answer extraction methods of our QA systems on each dataset. Both BERT-large-uncased and ALBERT-base-uncased use whole word masking and are pre-trained on SQuAD v1.1, while T5-large was pre-trained on \textit{Colossal Clean Crawled Corpus} or \textbf{C4}. Figure \ref{fig:t5_gt_answer_lengths} shows the comparison of generated answer lengths on the CovidGQA dataset.

\begin{table}[h!]
\centering
\begin{tabular}{||c c c c||} 
 \hline\hline
 Dataset & QA System & F1 Score & EM \\ [0.5ex] 
 \hline\hline
 & ALBERT-base & 23.37 & \textbf{13.04} \\ 
    CovidQA & BERT-large & \textbf{26.32} & 11.59 \\  
     & T5-large & 5.16 & 0.00 \\  
 \hline
 & ALBERT-base & 28.08 & \textbf{9.64} \\ 
    CovidGQA & BERT-large & \textbf{29.96} & 5.08 \\ 
     & T5-large & 5.95 & 0.00 \\ 
 \hline
\end{tabular}
\caption{ Preliminary results obtained using QA systems on CovidQA and CovidGQA.}
\label{table:1}
\end{table}

BERT-large achieves the highest macro-averaged F1 score on both datasets. Furthermore, BERT-large outperforms ALBERT-base on GLUE, RACE, and SQuAD benchmarks \cite{lan2020albert}. However, ALBERT-base unexpectedly outperforms BERT-large on EM, achieving the highest EM of the three models. Further experimentation is required to investigate the cause of this. Despite T5-large having the most significant number of parameters (770M), both BERT-large and ALBERT-base outperform T5-large on all metrics on both datasets. This may be explained by the fact that T5-large was not pre-trained on a QA task.
The reasonable performance of these three transformers, motivated us to use them to develop QA systems for the CORD-19 Kaggle competition. The next section discusses the QA systems in more detail.

\section{QA Systems for COVID-19}
Based on the results showed in the previous section, we have developed two QA systems on the CORD-19 dataset which aim to 
help the medical community answer high-priority scientific questions such as \textit{``What is the efficacy of novel therapeutics being tested currently?"} or \textit{``What is the best method to combat the hypercoagulable state seen in COVID-19?"}. We designed two BERT-based question-answering systems and a T5 question-answering system. 
Each QA system is pre-trained on different datasets and evaluated on two COVID-19 datasets.
\subsection{SBERT-BERT-QA} 
The SBERT-BERT-QA system (see Figure \ref{fig:Bert} \footnote{\footnotesize{The system diagram of SBERT-BERT-QA and LDA-ALBERT-QA. The dotted arrows and the white rectangles represent the path of SBERT-BERT-QA.}}) combines Sentence-BERT and BERT-large. We first filter the articles using a keyword search based on a set of pre-defined keywords such as \textbf{RNA virus, clinical, naproxen, clarithromycin}. Once filtered, the top $n$ articles were extracted by embedding the query and the article titles using a pre-trained Sentence-BERT model (i.e., a BERT-base model with mean-tokens trained on SNLI and MultiNLI corpora and then on STS Benchmark training set) \cite{reimers2019sbert}. Sentence-BERT (SBERT) is a modification of the BERT network using Siamese and triplet network structures to derive semantically meaningful sentence embeddings. The top $n$ articles were extracted by taking the articles with the highest cosine similarity scores between the query embedding and each article title embedding. Once the top article were extracted, each article's answer to the query was extracted using a pre-trained QA BERT-large-uncased model. 

\subsection{LDA-ALBERT-QA}
\begin{figure}[!tbp]
  \centering
  \begin{minipage}[b]{0.45\textwidth}
    \includegraphics[width=\textwidth]{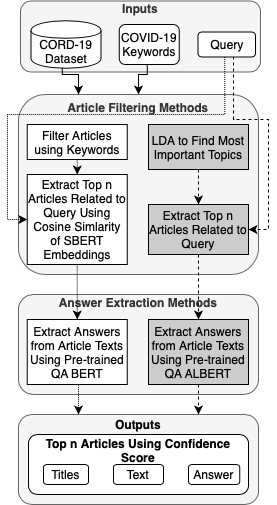}
    \caption{\footnotesize{BERT-based QAs}}
     \label{fig:Bert}
  \end{minipage}
  \hfill
  \begin{minipage}[b]{0.45\textwidth}
    \includegraphics[width=\textwidth]{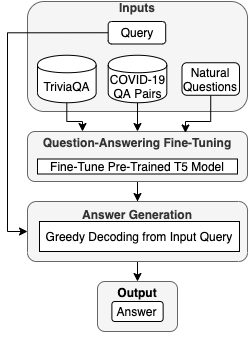}
    \caption{\footnotesize{Context-free T5 QA system}}
    \label{fig:T5}
  \end{minipage}
\end{figure}

The LDA-ALBERT-QA (see Figure \ref{fig:Bert}\footnote{ \footnotesize{The dashed arrows and the dark rectangles represent the path of LDA-ALBERT-QA.}})  combines LDA and pre-trained ALBERT-base. First, we filter out the irrelevant articles from the CORD-19 dataset using LDA to provide a dataset containing only the relevant articles to the query. Then, the filtered documents are fed into a pre-trained QA ALBERT-base-uncased model to extract an excerpt from articles that are relevant to the query. We have utilized \textbf{SQuAD v1.1} \cite{rajpurkar2016squad} and \textbf{BioASQ} 6b factoid QA pairs to develop the pre-trained ALBERT model. The primary reason for using SQuAD v1.1 is to overcome the data shortage of BioASQ 6b factoid, which contains less than 1k QA pairs. After pre-training the model on SQuAD v1.1, we further pre-train the model with BioASQ dataset to target biomedical domain. A single output layer on top of the pre-trained ALBERT model performs token-level classification to compute the start/end index of a predicted answer from each article.

\section{Conclusion}
We presented two transformer-based QA systems to compete in the competition. We selected transformers based on the preliminary results obtained from BERT, ALBERT, and T5 model on two QA datasets. As our results indicated, BERT-large could achieve the highest F1-score for both datasets and one of our first candidates to develop a QA system. Our results also showed that ALBERT-base could achieve the highest EM score for both datasets and was our second nominee to establish a QA system.      
We consider that one of the significant limitations of transformers is that they require a lot of labeled QA pairs to reach acceptable performance. We aimed to resolve the limitation by developing a hybrid QA system that combines few-shot learning with a transformer. LDA also has some limitations; for example, we need to determine the number of topics and consider the articles that are not too short. To address these issues, we explore prediction-focused supervised LDA and identify topics in an online manner.
Furthermore, a QA system should be explainable and trustworthy to medical users. Developing such a QA system is possible if medical experts and NLP researchers cooperate closely. 
\section*{Author's contributions}

The first three authors equally worked with technical parts of the paper that has been summarized in Section 3 and 4. Dr. Parsa wrote sections 1 and 2 and reviewed the manuscript and provided valuable feedback on the manuscript.  
\bibliographystyle{unsrt}  
\bibliography{references}  


\end{document}